\documentclass[10pt,twocolumn,letterpaper]{article}

\usepackage{cvpr}
\usepackage{times}
\usepackage{epsfig}
\usepackage{graphicx}
\usepackage{amsmath}
\usepackage{amssymb}
\usepackage{paralist}  

\usepackage[ruled,vlined]{algorithm2e}
\usepackage{lipsum}

\usepackage{subfig}
\usepackage{multirow}
\usepackage{array}
\let\mypdfximage\pdfximage
\def\pdfximage{\immediate\mypdfximage}

\newcommand{\vc}{{\mathbf c}}
\newcommand{\vd}{{\mathbf d}}
\newcommand{\vp}{{\mathbf p}}

\newcommand{\vx}{{\mathbf x}}

\cvprfinalcopy 
\def\cvprPaperID{1153} 

\ifcvprfinal\fi
\begin{document}

\title{ Efficient Feature Matching by Progressive Candidate Search }

\author{Sehyung Lee\\
Hanyang University\\
Seoul, Korea\\
{\tt\small shl@incorl.hanyang.ac.kr}
\and
Jongwoo Lim\\
Hanyang University\\
Seoul, Korea\\
{\tt\small jongwoo.lim@gmail.com}
\and
Il Hong Suh\\
Hanyang University\\
Seoul, Korea\\
{\tt\small ihsuh@hanyang.ac.kr}
}

\maketitle

\begin{abstract}
We present a novel feature matching algorithm that systematically utilizes the geometric properties of features such as position, scale, and orientation, in addition to the conventional descriptor vectors.
In challenging scenes with the presence of repetitive patterns or with a large viewpoint change, it is hard to find the correct correspondences using feature descriptors only, since the descriptor distances of the correct matches may not be the least among the candidates due to appearance changes.
Assuming that the layout of the nearby features does not changed much, we propose the bidirectional transfer measure to gauge the geometric consistency of a pair of feature correspondences.
The feature matching problem is formulated as a Markov random field (MRF) which uses descriptor distances and relative geometric similarities together.
The unmatched features are explicitly modeled in the MRF to minimize its negative impact.
For speed and stability, instead of solving the MRF on the entire features at once, we start with a small set of confident feature matches, and then progressively search the candidates in nearby features and expand the MRF with them.
Experimental comparisons show that the proposed algorithm finds better feature correspondences, i.e. more matches with higher inlier ratio, in many challenging scenes with much lower computational cost than the state-of-the-art algorithms.
\end{abstract}

\section{Introduction}

Local features are widely used in many applications including 3D scene reconstruction, image retrieval, image stitching, and object recognition. 
Among them, SIFT~\cite{lowe2004distinctive} and SURF~\cite{bay2006surf} are the most popular due to their invariance to illumination and small affine deformation.
The feature correspondences between two images are usually found by comparing the feature descriptors only.
The incorrect matches are filtered by RANSAC~\cite{fischler1981random}, and the inlier feature matches  are kept for further processing.
This framework has been successfully used in many applications, however, it often fails in challenging situations, such as the presence of repetitive patterns, or large illumination or viewing direction changes.

Most local features provide the scale and orientation in addition to the feature position, and this information has been used in feature matching. 
For example, when the feature descriptors are not discriminative enough, ambiguous match candidates can be removed by such additional information.
Several feature matching algorithms utilizing the additional pairwise information between neighboring features have been proposed, and achieved more robust feature matching than simple descriptor matching approach. 
One of the popular directions is the graph matching such as the spectral clustering~\cite{leordeanu2005spectral}, Monte-carlo sampling~\cite{suh2012graph}, and tensor decomposition \cite{duchenne2011tensor}.
Another direction is using the Markov random fields (MRF) such as dual decomposition~\cite{torresani2008feature,liu2014deformable} or graph cut~\cite{duchenne2011graph}.

In spite of success of these approaches, there are still several limitations. 
When there exist \emph{unpaired features}, the features which do not have true matches in the other image, the error by their incorrect match candidates is propagated through the whole graph, especially in MRF models, and affects the estimation of the other feature matches.
Another drawback of the graph optimization algorithms is their high computational cost.
The recent algorithms with less complexity~\cite{zhou2013deformable, zhou2012factorized, chen2013robust} are still not fast enough for solving large scale matching problems. 
These are the main reasons why most existing algorithms~\cite{duchenne2011tensor, cho2010reweighted, zass2008probabilistic, torresani2008feature, leordeanu2005spectral, suh2012graph} deal with a relatively sparse and small feature set of features.

\begin{figure*}
\centering
\includegraphics[width=\linewidth]{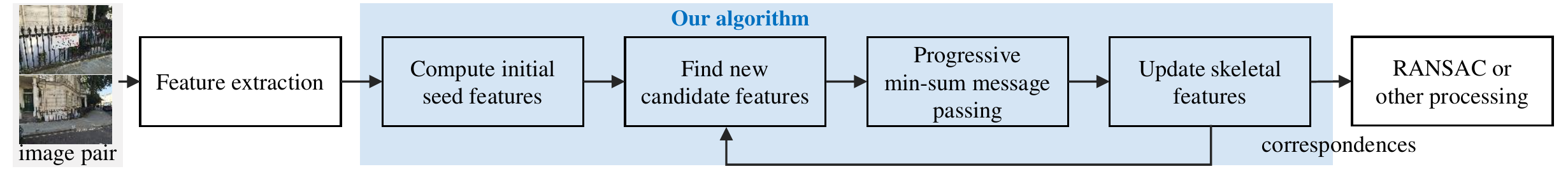}
\vspace{-0.7cm}
\caption{\small
Overview of the proposed algorithm.
}
\vspace{-0.3cm}
\label{fig:overview}
\end{figure*}

In this paper we propose a unified feature matching framework that considers both geometric and appearance properties of commonly-used local features (Figure~\ref{fig:overview}).
Similar to the MRF-based algorithms, the feature correspondences are found by solving an MRF on the feature set, whose unary and binary energy terms represent feature descriptor distances and relative geometric inconsistencies.

To mitigate negative impact of unpaired features, we introduce the `unmatched' label for each node in MRF with proper unary and binary costs.
Also the `bidirectional transfer measure' computes the relative geometric consistency between two feature correspondences more robustly.

The `progressive correspondence search' process, which identifies correspondences of more confident features first, then expands to other features, also helps reducing the chance of having incorrectly matched features in the graph.
Our progressive correspondence search also greatly reduces search time by starting with good initial matches at each stage.
In matching around two thousand features in each image, the proposed algorithm finds higher quality feature correspondences within 0.4 s, whereas other algorithms take several seconds to minutes.
Extensive experimental evaluation demonstrates that our approach effectively handles the limits of conventional approaches and achieves significant improvement in challenging feature matching problems.

The main contributions of our work are two fold.
\begin{compactitem}
\item The proposed MRF framework with the unmatched label and bidirectional transfer measure can handle the challenging scenes better and yields much improved feature matches.
\item Progressive construction and optimization of the MRF based on spatial neighborhood and multiple candidacy achieves higher accuracy, better robustness, and much faster processing.
\end{compactitem}

\section{Related Work}
\label{sec:related_work}

After SIFT \cite{lowe2004distinctive} and SURF \cite{bay2006surf}, many novel feature detection algorithms such as BRISK \cite{leutenegger2011brisk}, BRIEF \cite{calonder2012brief}, ORB \cite{rublee2011orb} and KAZE \cite{alcantarilla2012kaze}, were developed to improve SIFT's performance. 
In these algorithms, feature-to-feature matching with descriptors is the most common method for computing correspondences. 
In spite of the distinctiveness of the proposed descriptor, it is still insufficient for matching features in the presence of severe transformation.
To compensate for the limitations of simple descriptor matching, many algorithms have been proposed. 
We briefly review related works to clarify the differences between the proposed and existing methods.

To quickly find a good initial matches using feature descriptors, several heuristic algorithms have been developed.
The nearest neighbor and distance ratio (NNDR) algorithm~\cite{lowe2004distinctive} is based on the insight that the match is ambiguous if the best and the second-best match candidates of a feature have similar descriptors.
A match is accepted only if the ratio of the two descriptor distances is below a threshold.
NNDR effectively discards the ambiguous matches, but as a consequence the number of feature matches are often much smaller than the number of features.
Note that the following stages after feature matching, such as RANSAC, further filter inconsistent feature matches, thus it is critical to have sufficient number of good feature correspondences in the feature matching stage.

Our approach is related to several previous works.  
Leordeanu and Hebert~\cite{leordeanu2005spectral} introduced pairwise geometric information to build the affinity matrix.
Torresani \etal~\cite{torresani2008feature} designed a complex cost function based on the appearance and spatial proximities which can be efficiently optimized by dual decomposition. 
Cho \etal~\cite{cho2009feature} employed the agglomerative clustering algorithm instead of spectral clustering~\cite{leordeanu2005spectral}.
Their work introduced various solvers, reweighted random walks \cite{cho2010reweighted} and Monte-Carlo sampling \cite{suh2012graph}. 
In these algorithms, the affinity matrix is constructed by combining the descriptor similarity and Euclidean distance between features.
Zhou and Torre \cite{zhou2012factorized} factorized the affinity matrix as a product of smaller matrices. 
Because of this matrix factorization, there is no need to explicitly store the full-size affinity matrix.
Zaragoza \etal~\cite{zaragoza2014projective} proposed a technique to compute an as-projective-as-possible warping that aims for the feature locations to be globally projective. 
This method is developed for image stitching application in which the correspondence algorithm is based on the multiple-homography fitting using their previous work \cite{chin2012accelerated}.
In~\cite{ lin2014bilateral}, proposed by Lin \etal, starting with a set of initial matches, their algorithm computes an affine motion field between two images by exploiting that each match defines a local affine transformation.
Given the motion field, more correspondences are then recovered by finding the nearest neighboring match in descriptor space for each feature.

The proposed work differs from these algorithms in several ways. 
We directly measure the geometric dissimilarity between correspondences using the position, scale, and orientation provided by local features, instead of the distances between neighboring features and their corresponding features~\cite{leordeanu2005spectral}, the length and direction differences of correspondences \cite{torresani2008feature} or homography transformation using additional affine region detector~\cite{cho2009feature,cho2010reweighted,suh2012graph}. 
The proposed algorithm can be used with most available local features as long as the geometric properties are provided, not limited to a few specific local features.
Also, unlike the previous approaches~\cite{zhou2012factorized, duchenne2011tensor, cho2010reweighted, torresani2008feature, leordeanu2005spectral} that perform estimation on the entire feature set at once, our algorithm incrementally finds the geometrically reasonable matches in whole feature set from the initial set of features.
Because of this, the proposed algorithm can handle a large number of features and produce more accurate matches with minimal computational cost.


\section{Problem Formulation}

\begin{figure}
\centering
\includegraphics[width=\linewidth]{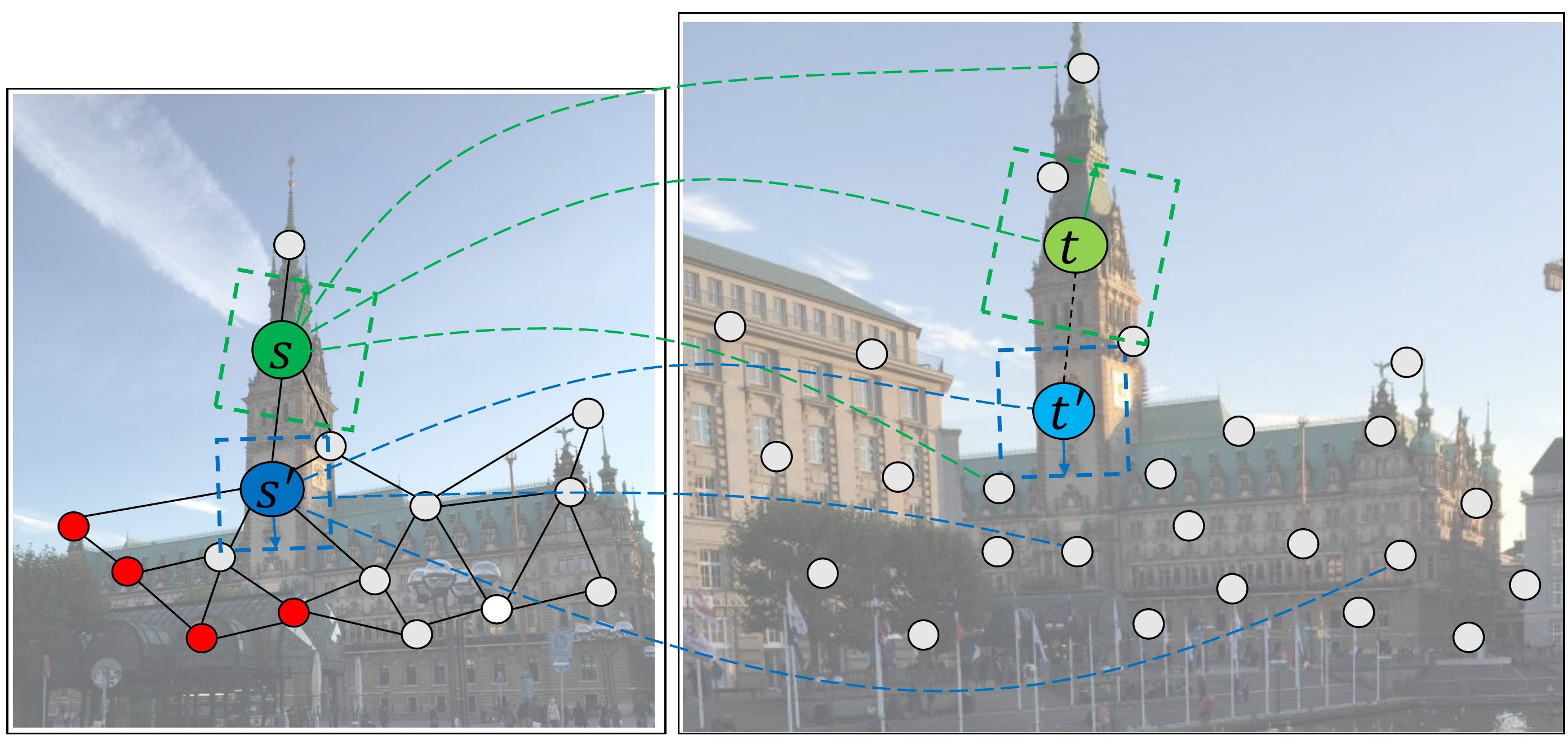}
\vspace{-0.5cm}
\caption{\small
The MRF model for feature correspondence.
The circles represent the detected features.
The MRF is constructed on the features in the reference image (left), and the solid lines in the reference image are the MRF edges.
Each node $i$ has its potential correspondence $c_i$ which is shown as dotted arcs between two views.
The pairwise energy between $(s_i,t_i)$ and $(s_j,t_j)$ affects the MRF cost, the red circles represent the features that their true correspondences do not exist in the target image.
}
\vspace{-0.3cm}
\label{fig:graphical_model}
\end{figure}

The feature matching task can be formulated as an energy minimization problem of the MRF constructed on the features. 
Given two sets of features detected in the reference and target images, the MRF model is constructed on the features in the reference image. 
The variable in each node represents its correspondence in the target image, and the entire cost of the MRF model represents the overall consistency of the current correspondences between two images. 
Figure~\ref{fig:graphical_model} visualizes the formulation.

The energy function of MRF is comprised of the unary and pairwise potentials.
The unary potential $E_\phi$ represents the appearance model to measure the descriptor dissimilarity of the current correspondence.
The pairwise potential $E_\psi$ models the relative geometric inconsistency between the pairs of correspondences. 
Then, the best correspondence set $C^*$ is obtained that minimizes the overall energy, which is written as
\vspace{-0.1cm}\begin{equation}
C^* = \underset{C}{\arg \min }  \ E_\phi(C) + \lambda E_\psi(C),
\label{eq:energy_function}
\vspace{-0.1cm}\end{equation}
where $C = \{ \vc_i \}$ is the set of feature correspondences, $\vc_i = (s_i, t_i)$, $s_i$ is the feature index in the reference image and $t_i$ is either the feature index in the target image or $\varnothing$ if $s_i$ is an unmatched/unpaired feature.
$\lambda$ is the weight parameter for leveraging the unary and pairwise models ($\lambda$ is set to 0.1).
To reduce the search space, for each feature $s$ we only consider the top-$\kappa$ descriptor matches in the target image as the candidate matches, i.e. $\vc_i\in\{(s_i,t_k)\}_{k=1\ldots\kappa}\cup\{(s_i,\varnothing)\}$ ($\kappa$ is set to 15).
Note that the size of the MRF graph is not affected by $\kappa$, but by the number of neighbor features.

The unary term $E_\phi$ measures the descriptor similarity between corresponding features. 
\vspace{-0.1cm}\begin{equation}
E_\phi(C) = \sum_i^n  e_\phi(\vc_i), \;\;\textrm{and}\;\; \\ \
\vspace{-0.1cm}\end{equation}
\vspace{-0.1cm}\begin{equation}
e_\phi(\vc_i) =  \left\{\begin{array}{cl}
\left\| \vd_{s_i} - \vd_{t_i} \right\| & \text{if $t_i\ne\varnothing$} \\ 
\alpha & \text{otherwise} \\
\end{array}\right.,
\label{eq:unary_potential}
\vspace{-0.1cm}\end{equation}
where $\vd_{s_i}$ and $\vd_{t_i}$ are the feature descriptor vectors of the features $s_i$ and $t_i$.
The unary energy term penalizes dissimilar feature correspondences, and 
the penalty term for unmatched features $\alpha$ is set to 0.5.

\begin{figure}
\centering
\includegraphics[width=\linewidth]{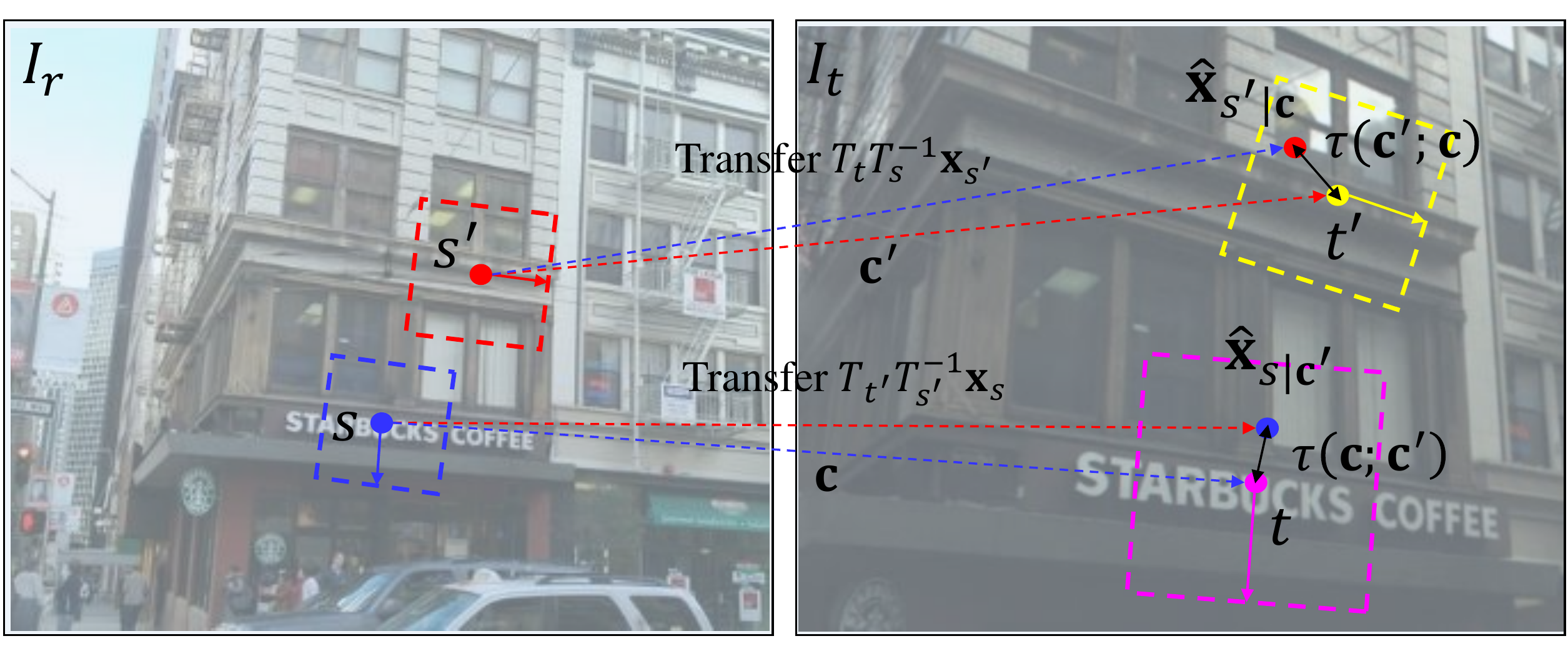}
\vspace{-0.7cm}
\caption{\small
The bidirectional transfer measure. 
The color points represent the detected local features. 
The arrows and dotted rectangles are the orientations and scales of the features. 
The neighboring feature is warped into the opposite image, and the transfer error is measured by the distance between original and transferred locations, $\hat\vx_{s'|\vc}$ and $\hat\vx_{t'}$. 
}
\vspace{-0.3cm}
\label{fig:bidirectional_transfer}
\end{figure}

{\flushleft \textbf{Bidirectional Transfer Measure. }}
In most local features the scale and orientation of each feature point is available as a by-product in realizing the scale and orientation invariance.
Although the scales, orientations, or positions of the features change according to camera pose or scene changes, their relative difference tends to remain similar if the features belong to the same object.
We propose the bidirectional transfer distances between neighboring features to measure the relative difference between their geometric properties.

Let $T_i$ be the transformation that maps a unit patch, which is centered at the origin, axis-aligned, and unit-sized, onto the patch of the feature $i$.
In case of SIFT or SURF, each feature patch is represented as x-, y-position, scale, and orientation $(tx_i, ty_i, \sigma_i, \theta_i)$, and then $T_i$ is written as a 2D similarity transform
\vspace{-0.1cm}\begin{equation*}
T_i = 
\begin{bmatrix}
\sigma_i \cos\theta_i & -\sigma_i \sin\theta_i & tx_i \\ 
\sigma_i \sin\theta_i & \sigma_i \cos\theta_i & ty_i \\ 
 0 & 0 & 1
\end{bmatrix}.
\vspace{-0.1cm}\end{equation*}
Depending on the geometric properties provided by local features, $T_i$ can be any type of 2D  transformation including similarity, affine or homography.

For a feature correspondences $\vc = (s,t)$, the warped position of a nearby feature $s'$ in the target image can be written as $\hat\vx_{s'|\vc} = T_{t} T_{s}^{-1} \vx_{s'}$.
If the relative geometry between two feature correspondences $\vc$ and $\vc'=(s',t')$ is approximately preserved between two images, the position of the matched feature in the target image $\vx_{t'}$ must be close to $\hat\vx_{s'|\vc}$.
Therefore the one-way transfer distance $\tau(\vc'; \vc)$ is then defined as
\vspace{-0.1cm}\begin{equation*}
\tau(\vc'; \vc) = \left\{\begin{array}{cl}
 \left \|\, \hat\vx_{s'|\vc} - \vx_{t'} \right \|^2, & \text{if $t\ne\varnothing$ and $t'\ne\varnothing$} \\ 
 0 & \text{otherwise} \end{array}\right.,
\vspace{-0.1cm}\end{equation*}
and the pairwise energy $e_\psi(\vc, \vc')$ is defined as
\vspace{-0.1cm}\begin{equation*}
e_\psi(\vc, \vc') = \tau(\vc'; \vc) + \tau(\vc; \vc') +
  \tau(\vc'^{-1}; \vc^{-1}) + \tau(\vc^{-1}; \vc'^{-1}),
\vspace{-0.1cm}\end{equation*}
where $\vc^{-1} = (t,s)$ represents the inverse correspondence of $\vc$ defined by flipping the reference and target features.
It considers the differences not only by the forward warping but also by the backward warping.
and the pairwise energy $E_\psi$ of the MRF is as follows:
\vspace{-0.1cm}\begin{equation*}
\ E_\psi(C) = \sum_i^n \sum_{j \in N(i) }^m e_\psi(\vc_i, \vc_j),
\label{eq:pairwise_potential}
\vspace{-0.1cm}\end{equation*}
where $N(i)$ is the set of neighbors of the $i$-th feature.

Our bidirectional transfer model is based on the similarity transform obtained by the relative changes of potential matches. 
When affine-based features such as ~\cite{mikolajczyk2004scale,morel2009asift} are used, the bidirectional transfer model can be easily extended to the affine transformation. 
However, they are much slower than other widely used algorithms \cite{leutenegger2011brisk, alcantarilla2012kaze, rublee2011orb, calonder2012brief, lowe2004distinctive, bay2006surf}, and in this paper we only use the similarity model.

{\flushleft \textbf{Belief Propagation. }}
The problem of Equation~\ref{eq:energy_function} is NP-hard, thus in general it is not possible to find the optimal solution.
Instead we employ the belief propagation (BP) algorithm to find the approximate solution. 
Among the many variants of BP algorithms, the min-sum message passing is used due to its computational efficiency compared to sum-product or max-product methods. 
The message from the node $i$ to the node $j$ for the correspondence $\vc_j$ is
\vspace{-0.1cm}\begin{eqnarray*}
m_{i\rightarrow j}(\vc_j) = 
\underset{\vc_i}{\min} \Big[e_\phi(\vc_i) + \lambda e_\psi(\vc_i, \vc_j) + \hspace{-0.3cm}\sum _{k\in N(i) \setminus j}\hspace{-0.3cm} m_{k\rightarrow i}(\vc_i) \Big].
\label{eq:message_passing}
\vspace{-0.1cm}\end{eqnarray*}
After convergence, the belief of a correspondence $\vc_i$ in each node $i$ is computed as 
\vspace{-0.1cm}\begin{equation}
b(\vc_i) = e_\phi(\vc_i) + \sum_{k\in N(i)} m_{k\rightarrow i}(\vc_i). 
\vspace{-0.1cm}\end{equation}
The optimal feature correspondence set $C^*$ is determined by individually selecting the label having minimum cost in each node.

\section{Progressive Feature Matching}

It is possible to build an MRF with all detected features at once and try to find the solution of Equation~\ref{eq:energy_function} using BP, but it often converges to a wrong solution and takes long time to converge due to the large amount of incorrect or confusing initial feature correspondences.
To reduce this risk and achieve faster convergence, we propose a progressive feature matching scheme.
Once the seed features with good correspondences are available, then they can guide the matching process of the nearby features from similar part of the scene.
This motivates the progressive search of feature correspondence.
Starting from a small set of highly confident candidate matches, the BP algorithm finds the converged correspondence assignments, and the unmatched features are excluded from the MRF.
Then the MRF is iteratively expanded by nearby features around the existing features, followed by BP optimization and filtering, until all features are considered.

%
{\flushleft \textbf{Initial Seed Features.}}
To reduce the chance of selecting unpaired features, we select the features with high descriptor similarity scores as the initial seed features.
Among the features in the reference image whose NNDR ratio is smaller than $\theta$, the top-$r$ features in terms of the descriptor score are used as the nodes of the initial MRF model, and each node is connected with its $K$-nearest neighbors (we use $\theta=0.9$, $r=100$, and $K=5$ in our experiments).
%
Although we selected high-quality matches for the initial MRF, after the BP process, there may exist unmatched features (the cost of `unmatched' is the smallest).
These features are excluded from the MRF graph and put back to the candidate feature set.

\begin{figure}
\centering
\includegraphics[width=\linewidth]{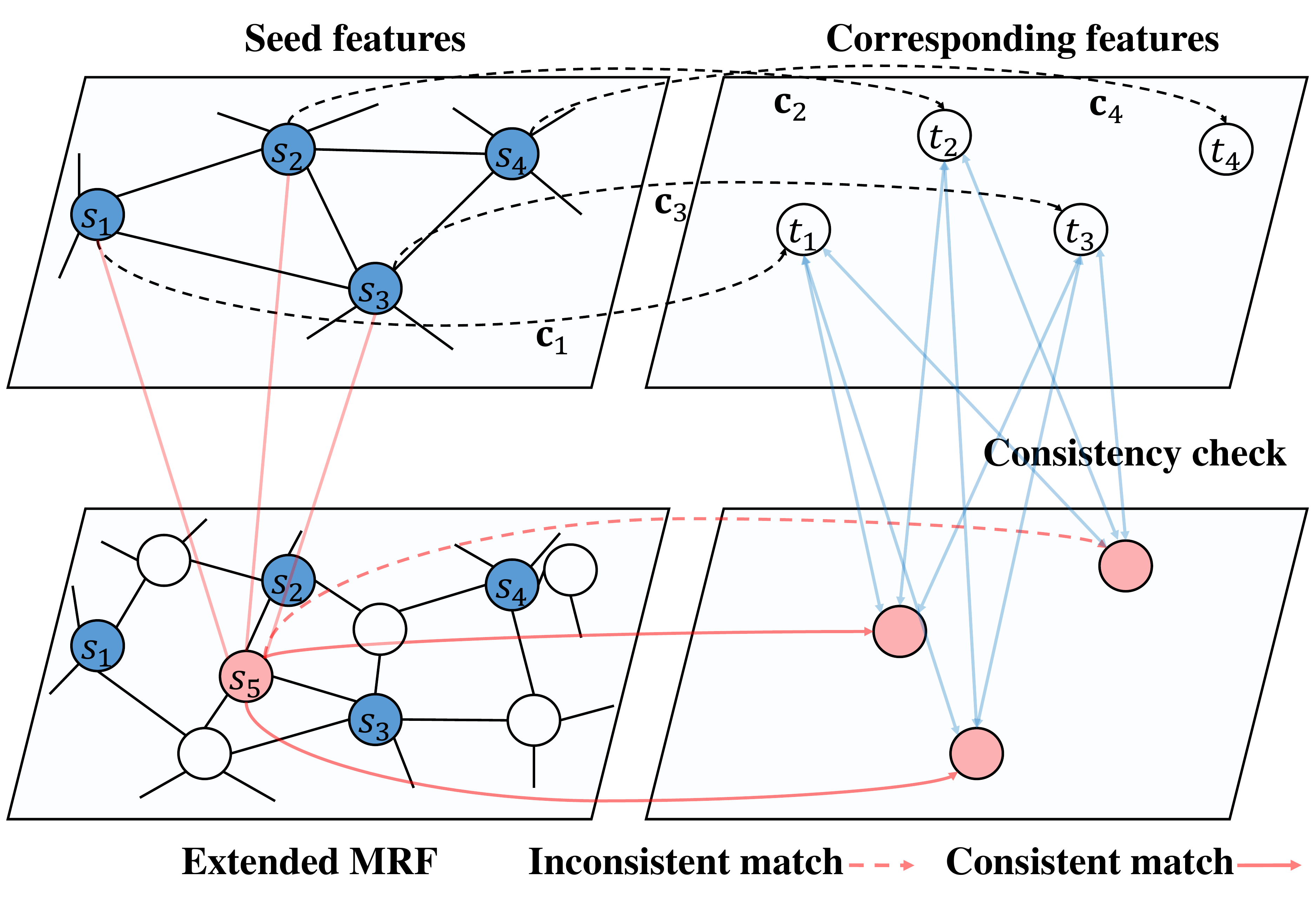}
\vspace{-0.7cm}
\caption{ \small
Overview of progressive feature matching.
After the correspondences of seed features (blue circles) and their correspondences are computed (black dotted lines), our MRF model is expanded. 
By comparing the candidate matches of the extended features with the nearby seed correspondences, only geometrically consistent matches are considered as candidate matches in the expanded MRF model.
}
\vspace{-0.3cm}
\label{fig:progressive}
\end{figure}

{\flushleft \textbf{Progressive Candidate Search.}}
The features in the MRF of the previous iteration are called as the `seed' features in the current iteration.
In our framework, their feature correspondences are used as the guides to find the correspondences of neighboring features when there are multiple candidate matches with similar descriptor scores.
Among all features not in the MRF, the $K$-nearest neighbor (KNN) features of the `seed' features are selected and added to the MRF as the `extended' nodes.
The KNN nodes of each extended nodes are connected with edges.
In the progressive iteration, the correspondences of the seed features are kept fixed, and the seed nodes only emit the messages toward the neighboring extended nodes in the BP process.
By comparing the candidate matches of the extended features with the nearby seed correspondences, geometrically inconsistent matches are quickly excluded in the message passing. 
For speed-up, the inconsistent match candidates with any of nearby seed nodes are explicitly excluded from the consideration.
By fixing the seed correspondences and reducing the number of candidate matches of extended nodes, it is possible to keep the computational complexity in a reasonable level.
The above-mentioned process is described in Algorithm~\ref{algorithm:progressive_min_sum} and illustrated in Figure~\ref{fig:progressive}.
We provide detailed description of the algorithm in the next paragraphs.

Given the seed set $S = \{s_i\}$ and their correspondences $C_S = \{\vp_i = (s_i, t_i)\}$, every leftover feature $f\notin S$ is tested if it has any nearby seed correspondence that has similar relative motion, i.e.
\vspace{-0.1cm}\begin{equation}
min_{l=1\ldots\kappa}\left(min_{k\in N_S(f)} (e_\psi((f,g_l), \vp_k))\right) < \theta_{seed},
\label{eqn:consistency}
\vspace{-0.1cm}\end{equation}
where $N_S(f)$ is the KNN seed features of $f$, $g_l$ is the index of top-$\kappa$ candidate matches of $f$, and $\theta_{seed}$ is set to 80.
The extended feature set $Q$ contains the features that satisfy the above condition.
The MRF is built upon $S\cup Q$ by connecting each node in $Q$ with its KNN nodes.
Note that the correspondence of the seed nodes are fixed, thus it is not necessary to perform any message passing to the seed nodes.
Only the messages toward the extended nodes are computed and updated.

In the MRF of the initial seed features, all $\kappa$ correspondence candidates are considered, but in the progressive candidate search, only the candidates that are consistent with any of the connected seed nodes (Equation~\ref{eqn:consistency}) are involved in BP process.
This reduces the size of the message $m_{i\rightarrow j}$ drastically from $\kappa^2$ to $\kappa_i\times\kappa_j$ where $\kappa_i$ and $\kappa_j$ are the number of candidates of node $i$ and $j$ (including unmatched $\varnothing$) respectively.
If a node is a seed node, it only has one candidate, and in practice most extended nodes have around 5 candidates.
The min-sum BP algorithm is then run on the MRF to find the best feature correspondences of the extended features, except the unmatched features (the cost of `unmatched' is less than those of any candidate matches).
Since the time complexity of the BP algorithm scales linearly to the message size, the reduction from $\kappa^2$ to $\kappa_i\times\kappa_j$ ($\kappa_i, \kappa_j \ll \kappa$) greatly improves the running speed.

\begin{algorithm}[tb]
\small
\SetAlgoLined
Input: seed nodes $S$, seed correspondences $C_S=\{\vp_i\}$. \\
Output: extended correspondences $C_{S\cup Q}$. \\
$Q \leftarrow \{\}$  \ \ \ \ ... the extended feature set.\\
\ForEach{ $f$ }{
  \eIf{ $f \notin S$ }{
      Find $K$-nearest seed nodes $N_S(f)$. \\
    $\gamma_f \leftarrow \{\varnothing\}$ \ \ \ \ ... candidate matchs of $f$ to consider.\\
    \ForEach{ candidate match	 $(f,g_k)_{k=1\ldots\kappa}$ of $f$ }{
      \If{$min_{n\in N_S(f)} (e_\psi((f,g_k), \vp_n)) < \theta_{seed}$}{
        $Q \leftarrow Q \cup \{f\}$ \\
        $\gamma_f \leftarrow \gamma_f \cup \{g_k\}$ \\
      }
    }
  }{
  	Add seed feature and its correspondence. \\
	  $Q \leftarrow Q \cup \{s_f\}$   \\
      $\gamma_f \leftarrow \gamma_f \cup \{t_f\}$ \\
  }
}
Build the MRF using $Q$.\\
\Repeat{ convergence }{
  \ForEach{ extended node $f\in Q$ }{
    \ForEach{ $K$-nearest neighbor node $n$ of $f$ }{
      \ForEach{ $\vc = (f,g)$, $g \in \gamma_f$ }{
        $ m_{n\rightarrow f}(\vc) \leftarrow \underset{\vc_n=(n,*)}{\min} (  e_\phi(\vc_n) + \lambda e_\psi(\vc_n, \vc) + \sum _{l\in N(n) \setminus f} m_{l\rightarrow n}(\vc_n) ) $ \\
      }
    }
  }
  \ForEach{ extended node $f\in Q$ }{
    \ForEach{ $\vc = (f,g)$, $g \in \gamma_f$ }{
      $ b(\vc) \leftarrow e_\phi(\vc) + \sum_{l\in N(f)} m_{l\rightarrow f}(\vc)  $ \\
    }
  }
}
$C_{S\cup Q} \leftarrow C_S$ \\
\ForEach{ extended node $f\in Q$ }{
  \If{ $\arg\min_{\vc=(f,*)}(b(\vc)) \ne (f,\varnothing)$ }{
    $C_{S\cup Q} \leftarrow C_{S\cup Q} \cup \arg\min_{\vc=(f,*)}(b(\vc))$ \\
  }
}
 \caption{Progressive Min-Sum Message Passing }
\label{algorithm:progressive_min_sum}
\end{algorithm}

{\flushleft \textbf{Overall algorithm flow.}}
The proposed algorithm proceeds as follow:
\begin{compactitem}
\item Compute the initial seed features by selecting the features highly ranked in descriptor matching. 
\item Compute the correspondences of the seed features using our MRF model. 
\item Compute new child features and their possible matching hypotheses by comparing them with their parent features. 
\item Compute the correspondences of the child features through algorithm \ref{algorithm:progressive_min_sum}. 
\item Update the parent features, and iterate Steps 3 to 5 until new candidates are not found.
\end{compactitem}
Once the feature correspondences are computed, further post-processing such as RANSAC can follow as a conventional approach.

\begin{figure*}[t]
\centering
\includegraphics[width=\linewidth]{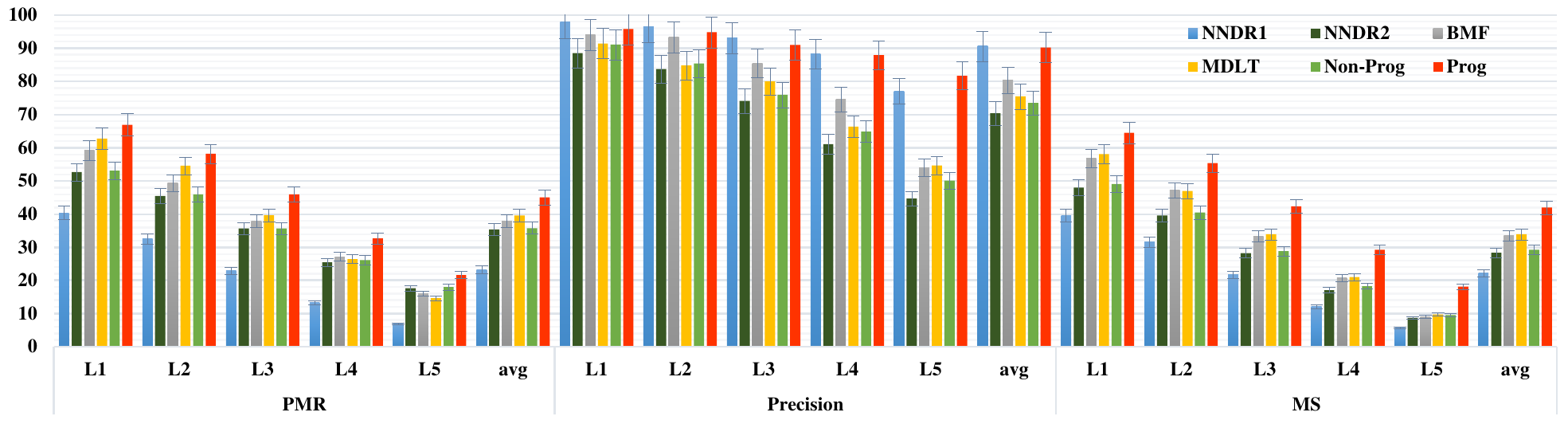}
\vspace{-0.8cm}
\caption{\small
Comparisons of feature matching algorithms (unit: \%).
It can be easily verified that the proposed algorithm outperforms the rest by a large margin.
}
\vspace{-0.3cm}
\label{fig:performance_comparison}
\end{figure*}

\section{Experimental Results}
We conducted extensive experimental evaluation of the proposed algorithm on various datasets.
The first set of experiments shows the quantitative comparison of the simple NNDR-based algorthms, the state-of-the-art methods \cite{zaragoza2014projective, lin2014bilateral}, and the proposed algorithm.
In the second set of experiments the improvement by the proposed method on several popular local features. 
We urge readers to see the supplementary material for complete experimental results and detailed explanations.
All experiments are conducted in MATLAB on a computer with i7 4.0 GHz CPU and 32GB memory, and our algorithm is partially implemented as MEX functions.
The source codes and dataset will be provided to public.

Heinly \etal \cite{heinly2012comparative} proposes three different quantitative evaluation metrics: \textit{putative match ratio}, \textit{precision}, and \textit{matching score}. 
The putative match ratio, PMR = $\frac{\#\text{putative matches}}{\#\text{features}}$, the ratio of computed matches to all detections, represents the selectivity of the matchers.
A less restrictive matching algorithm yields higher putative match ratios, whereas very restrictive matcher may discard potentially valid matches and end up with a low putative match ratio. 
The precision, Precision = $\frac{\#\text{inlier matches}}{\#\text{putative matches}}$, denotes the ratio of correct matches out of the putative matches (the inlier ratio). 
It measures the `quality' of feature matching, i.e. how many are correct among the declared feature matches.
For example, a very restrictive matcher may have a low putative match ratio but the precision can be high if most incorrect matches are filtered.
The matching score, MS = $\frac{\#\text{inlier matches}}{\#\text{features}}$, which is equivalent to PMR $\times$ Precision, show how many true matches are selected among all feature detections.
We can now quantify how many features were matched (PMR), and how many are true matches (Precision and MS).

\begin{table}[t]
\setlength{\tabcolsep}{6.0pt}
\renewcommand{\arraystretch}{1.0}
\begin{center}
\footnotesize
\begin{tabular}{ccccccccc}
\hline
PMR & L1 & L2 & L3 & L4 & L5 & avg  \\
\hline
NNDR1 & 40.39 &	32.53 &	23.00 &	13.34 &	6.92 &	23.24 \\
NNDR2 &	52.71 &	45.44 &	35.68 &	25.48 &	17.61 &	35.38 \\
BMF & 59.22 &	49.41 &	37.97 &	27.16 &	16.07 &	37.96 \\
MDLT & 62.78 &	54.55 &	39.68 &	26.45 &	14.59 &	39.61 \\
Non-Prog & 53.11 &	45.98 &	35.71 &	26.15 &	18.01 &	35.79 \\
Prog & 66.88 &	58.16 &	45.98 &	32.63 &	21.67 &	45.07  \\
\hline
\hline
Precision & L1 & L2 & L3 & L4 & L5 & avg  \\
\hline
NNDR1 & 97.91 &	96.58 &	93.06 &	88.29 &	77.13 &	90.60  \\
NNDR2 &	88.56 &	83.71 &	74.07 &	61.08 &	44.70 &	70.42  \\
BMF & 94.00 &	93.34 &	85.50 &	74.57 &	54.07 &	80.30  \\
MDLT & 91.44 &	84.85 &	79.99 &	66.41 &	54.72 &	75.48  \\
Non-Prog & 91.17 &	85.43 &	75.98 &	64.95 &	50.07 &	73.52  \\
Prog & 95.81 &	94.80 &	91.02 &	87.94 &	81.75 &	90.26   \\
\hline
\hline
MS & L1 & L2 & L3 & L4 & L5 & avg  \\
\hline
NNDR1 & 39.62 &	31.61 &	21.77 &	12.18 &	5.74 &	22.18  \\
NNDR2 &	47.95 &	39.59 &	28.25 &	17.06 &	8.78 &	28.32   \\
BMF & 56.80 &	47.20 &	33.37 &	20.75 &	9.09 &	33.44   \\
MDLT & 58.06 &	47.01 &	33.94 &	21.00 &	9.77 &	33.96   \\
Non-Prog & 49.09 &	40.43 &	28.81 &	18.26 &	9.56 &	29.23   \\
Prog & 64.52 &	55.36 &	42.33 &	29.29 &	18.12 &	41.92    \\
\hline
\end{tabular}
\end{center}
\vspace{-0.5cm}
\caption{\small
PMR, Precision, and MS of BMF~\cite{ lin2014bilateral}, MDLT~\cite{zaragoza2014projective}, non-progressive (Non-Prog), and progressive (Prog) methods on the test dataset (unit: \%).
The dataset~\cite{mikolajczyk2005comparison} contains 5 different levels of geometric and photometric transformation. 
From left to right, the average performance of each level (L1-L5), and total average are shown.
}
\vspace{-0.3cm}
\label{tab:comparison_table}
\end{table}

\subsection{Comparison to Existing Algorithms}
The proposed method is compared with three algorithms, NNDR with two different thresholds (0.9 for NNDR1, and 0.8 for NNDR2), MDLT~\cite{zaragoza2014projective}, and BMF~\cite{lin2014bilateral} using their implementations\footnote{BMF \cite{lin2014bilateral}: http://mmcheng.net/bfun/ \\ MDLT \cite{zaragoza2014projective}: https://cs.adelaide.edu.au/~jzaragoza/doku.php?id=mdlt}. 
In this experiment ASIFT~\cite{morel2009asift} feature is used since BMF only works with ASIFT.
Evaluation with other features are discussed in the next subsection.

\begin{figure}
\centering
\subfloat{\includegraphics[trim=0.5cm 0.5cm 0.5cm 0.5cm, width=0.95\linewidth]{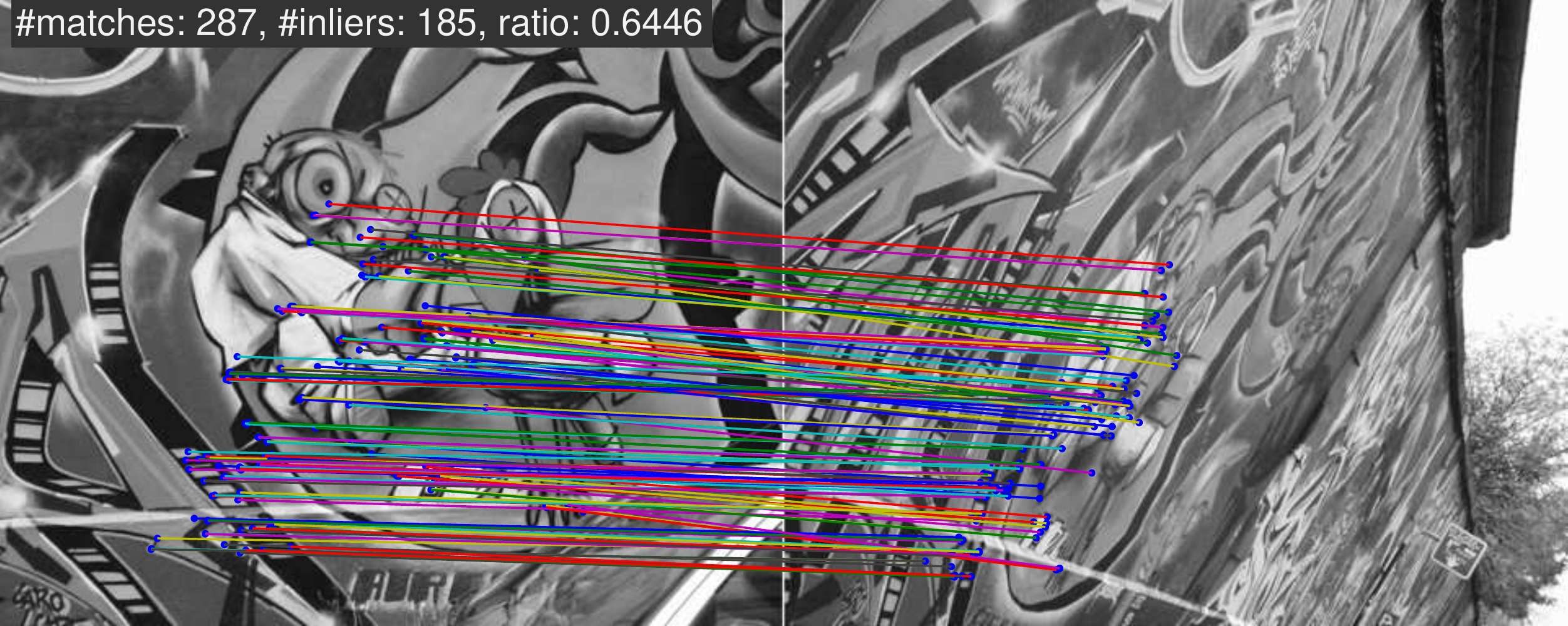}} \\
\subfloat{\includegraphics[trim=0.5cm 0.5cm 0.5cm 0.5cm, width=0.95\linewidth]{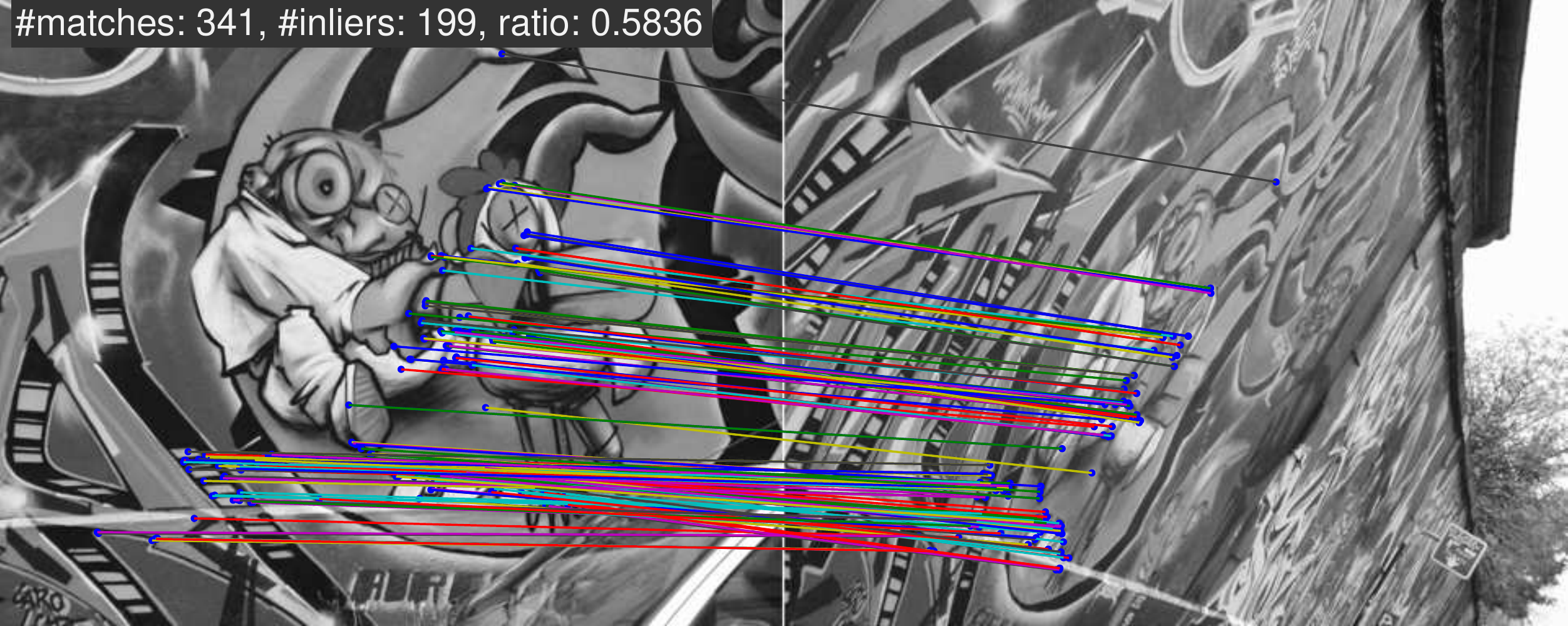}} \\
\subfloat{\includegraphics[trim=0.5cm 0.5cm 0.5cm 0.5cm, width=0.95\linewidth]{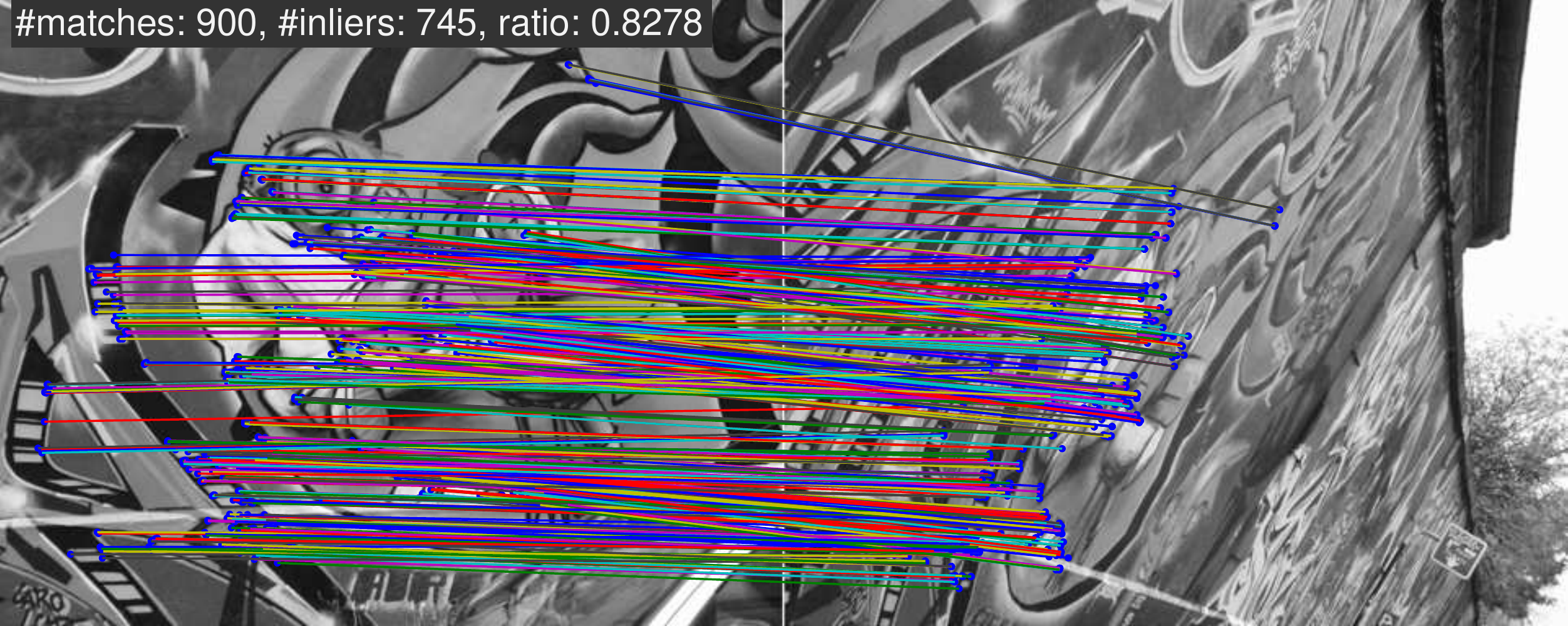}} 
\caption{ \small
From top to bottom, the feature matching results of BMF, MDLT, and Prog on the L5 pair of `graf' in the dataset~\cite{zaragoza2014projective}. The proposed algorithm finds more and better feature matches than others.
}
\vspace{-0.3cm}
\label{fig:correspondences2}
\end{figure}

For quantitative comparison, the well-known dataset constructed by Mikolajczyk~\etal \cite{mikolajczyk2005comparison} is used since it provides the ground-truth homography transformation.
It consists of 8 sets of 6 images, one reference and 5 target images, whose geometric and/or photometric properties are gradually changed.
The label from L1 to L5 denotes the amount of deformation or degradation.
The feature matches are considered as correct if the distance to the warped point by the ground-truth homography is smaller than 10 pixels.

Figure~\ref{fig:performance_comparison} and Table~\ref{tab:comparison_table} show the averaged performance in each level (L1-L5) and the average over all levels.
The proposed method (Prog) outperforms BMF and MDLT by a large margin in PMR and MS metrics in all levels, and in Precision it performs slightly better than the others.
We also test another version of our algorithm (Non-Prog) which runs without the progressive correspondence search to show the effectiveness of our progressive algorithm. 

In many tasks using features, having more inliers is as important as having high inlier ratio.
For example, in structure-from-motion, the inlier ratio determines the number of RANSAC iterations and the number of inliers affects the quality and fidelity of the 3D reconstruction.
The benefit of utilizing relative geometric information becomes significant when there exists a large deformation (in L5 or L4) - all metrics including Precision of the proposed method are much higher.
Figure~\ref{fig:correspondences2} shows a representative example of feature matching by different algorithms. 

Also in terms of the processing time, the proposed algorithm is much faster than the other algorithms.
The processing times of BMF, MDLT, Non-Prog and Prog on 1115 and 1078 SIFT features are 0.54, 4.82, 0.62, and 0.21 seconds, respectively. 
In the test with more number of features (2430 and 2196), the processing times are 3.92, 9.81, 7.27, and 0.36 seconds, respectively. 
The proposed algorithm is more efficient and effective than conventional feature matching algorithms and the non-progressive version.

\subsection{Experiments with Various Local Features}
Another advantage of the proposed method is its generality - it can be used with any local features. 
In this experiment, SIFT~\cite{lowe2004distinctive}, SURF~\cite{bay2006surf}, and KAZE~\cite{alcantarilla2012kaze} features in OpenCV 3.0 are tested. 
We compare the proposed method with NNDR of three different thresholds: 1.0 (Nearest), 0.9 (NNDR1), and 0.8 (NNDR2). 
Similar to the previous experiments, PMR, Precision, and MS of the feature matching results are compared.

\begin{figure}
\centering
\subfloat{\includegraphics[trim=0.5cm 0.5cm 0.5cm 0.5cm, width=0.95\linewidth]{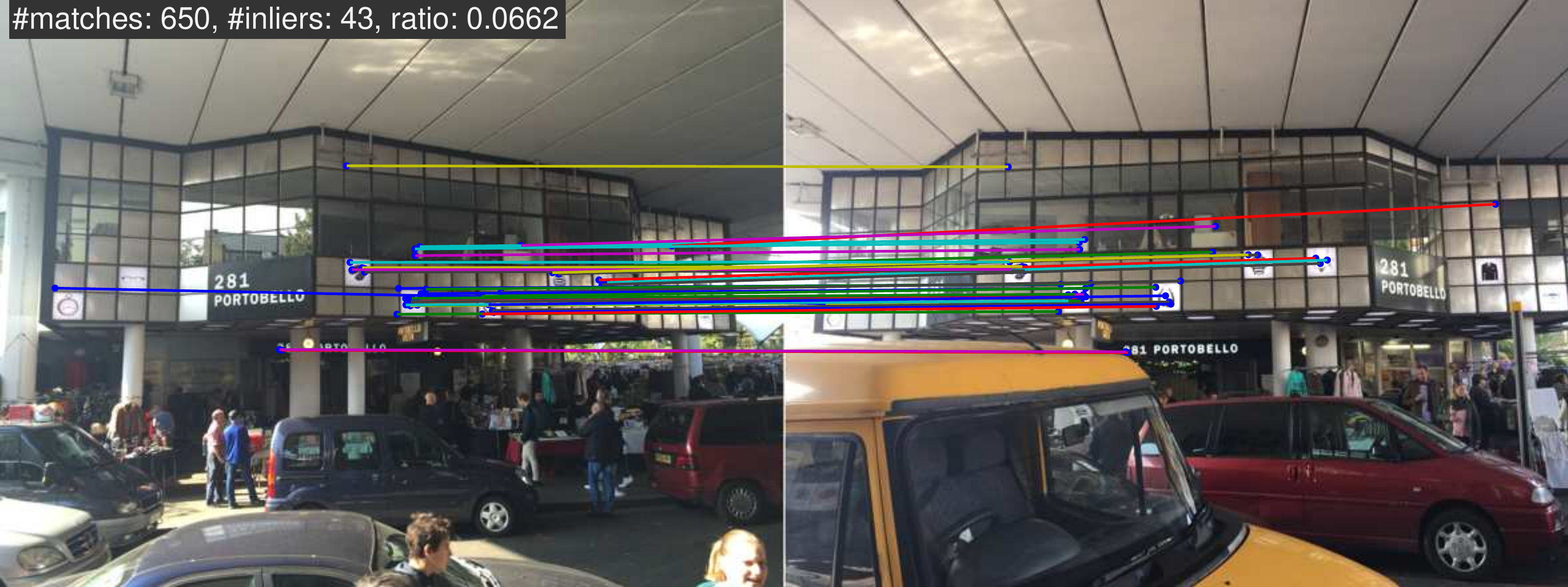}} \\
\subfloat{\includegraphics[trim=0.5cm 0.5cm 0.5cm 0.5cm, width=0.95\linewidth]{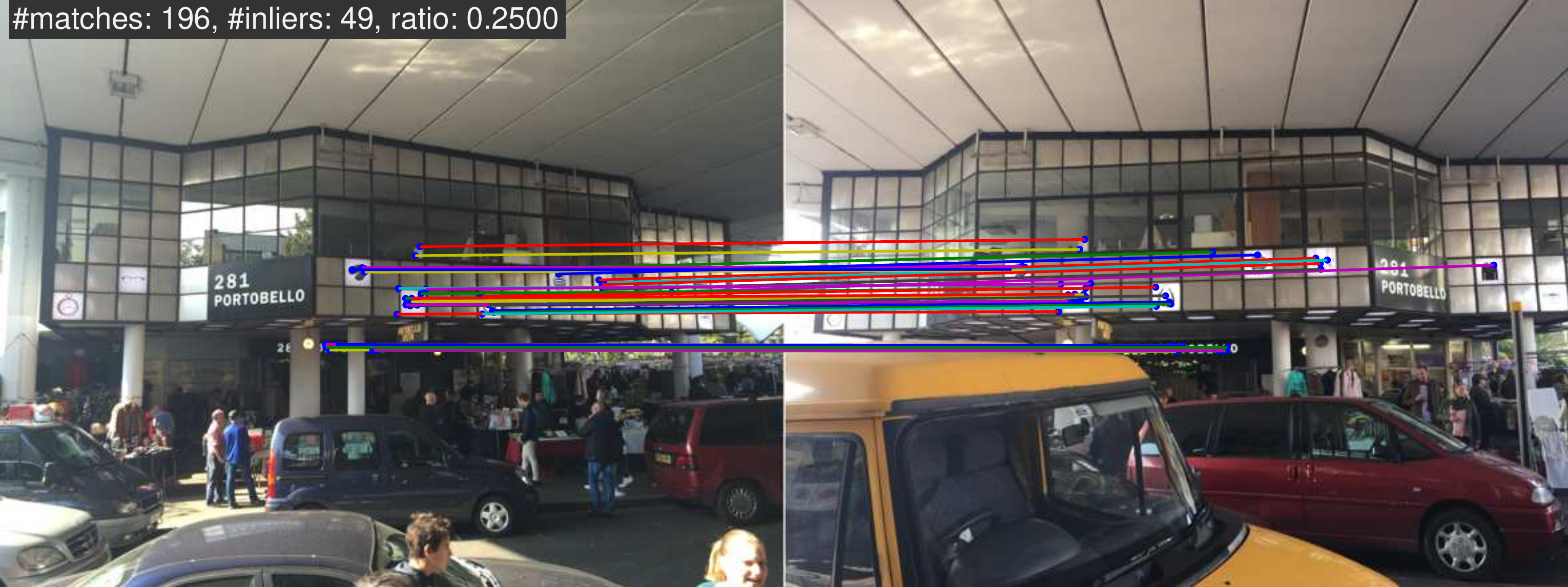}} \\
\subfloat{\includegraphics[trim=0.5cm 0.5cm 0.5cm 0.5cm, width=0.95\linewidth]{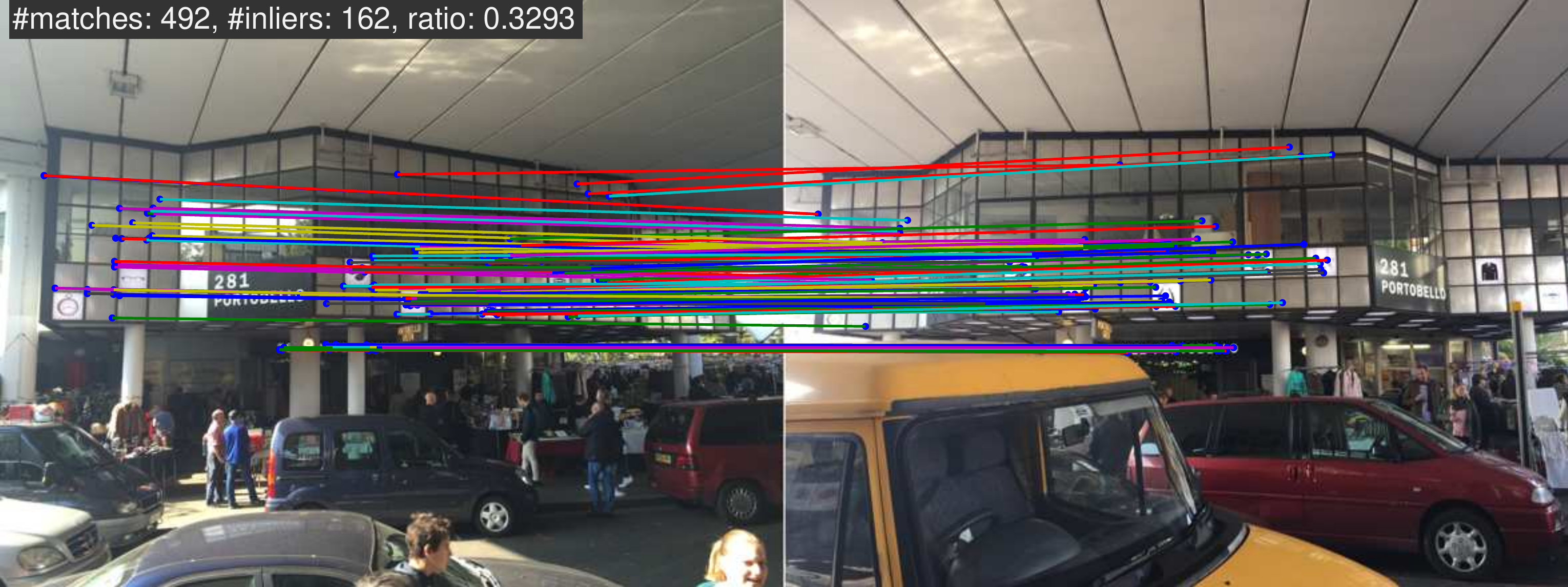}} 
\caption{ \small
The feature correspondence results on SIFT features. 
Each set of images show the results of NNDR1, NNDR2, and our method. 
The proposed algorithm produced much more inlier correspondences. 
Simple matching methods fail to find good matches due to many repetitive patterns and large viewpoint change. 
}
\label{fig:correspondences}
\end{figure}

In addition to Heinly's dataset~\cite{heinly2012comparative}, we constructed an extended dataset of 40 images with planar and non-planar scenes with large geometric and photometric variations.
Especially, the image pairs containing many repetitive patterns such as windows and decorative elements are included.
For quantitative evaluation, the inlier matches are found using the RANSAC algorithm with 8-point algorithm~\cite{hartley1997defense}, followed by manual elimination of a few obvious outliers.
Figure~\ref{fig:correspondences} displays one of the test images and the matching results using SIFT features. 
As shown in these images, our method produces significantly better correspondences with higher precision  (162 inliers and 32.93\% precision vs. 49 inliers and 25\% precision of NNDR2).
We encourage readers to see more matching results in the supplementary material.

\begin{figure}
\centering
\includegraphics[width=\linewidth]{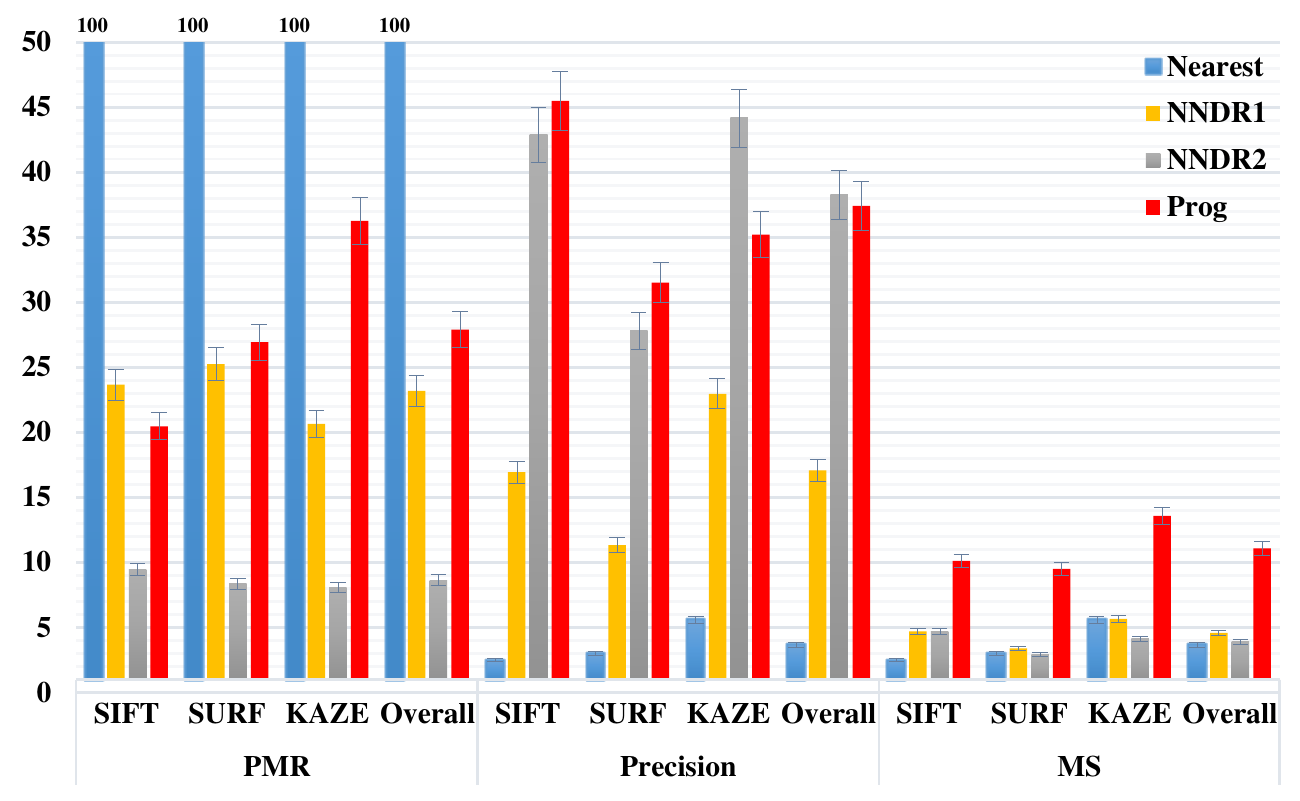}
\vspace{-0.7cm}
\caption{ \small
The feature matching performance on various local features: SIFT \cite{lowe2004distinctive}, SURF \cite{bay2006surf}, and KAZE \cite{alcantarilla2012kaze}  (unit: \%). 
The proposed method and NNDR with 3 different thresholds are compared.
Refer to the text for more detail.
}
\vspace{-0.3cm}
\label{fig:performance_comparison2}
\end{figure}

\begin{table}
\setlength{\tabcolsep}{8.5pt}
\renewcommand{\arraystretch}{1.0}
\begin{center}
\footnotesize
\begin{tabular}{ccccccccc}
\hline
SIFT & Nearest & NNDR1 & NNDR2 & Prog  \\
\hline
PMR & 100  & 23.67  & 9.49  & 20.49    \\ 
Precision & 2.52  & 16.94  & 42.91  & 45.49     \\ 
MS & 2.52  & 4.72  & 4.70  & 10.13       \\
\hline
\hline
SURF & Nearest & NNDR1 & NNDR2 & Prog  \\
\hline
PMR & 100 & 25.25  & 8.36  & 26.95    \\ 
Precision & 3.01  & 11.36  & 27.81   & 31.54    \\ 
MS & 3.01  & 3.40  & 2.93   & 9.51       \\
\hline
\hline
KAZE & Nearest & NNDR1 & NNDR2 & Prog  \\
\hline
PMR & 100 & 20.67  & 8.12  & 36.29    \\ 
Precision & 5.60  & 22.99  & 44.18  & 35.23     \\ 
MS & 5.60  & 5.68  & 4.13   & 13.60       \\
\hline
\hline
Overall & Nearest & NNDR1 & NNDR2 & Prog  \\
\hline
PMR & 100 & 23.20  & 8.65  & 27.91    \\ 
Precision & 3.71  & 17.10  & 38.30  & 37.42     \\ 
MS & 3.71  & 4.60  & 3.92   & 11.08       \\
\hline
\end{tabular}
\end{center}
\vspace{-0.5cm}
\caption{ \small
The feature matching performance on various local features: SIFT \cite{lowe2004distinctive}, SURF \cite{bay2006surf}, and KAZE \cite{alcantarilla2012kaze}  (unit: \%). 
The proposed method and 3 basic matching criteria are used for comparison.
The quantitative evaluation is based on averages of PMR, Precision, and MS metrics. 
}
\vspace{-0.3cm}
\label{comparison_table}
\end{table}

Figure~\ref{fig:performance_comparison2} summarizes the experimental results (averaged over the 40 image pairs). 
One can see significant performance improvement by the proposed algorithm compared to basic approaches. 
The proposed algorithm achieves the best MS in all local features, finding 2-4 times more inlier matches compared to any of NNDR-based methods.
%
In terms of precision, the proposed method shows similar performance to NNDR2.
Considering that NNDR2 yields very low PMR, which means it only accepts the matches with very distinct descriptors, the proposed algorithm achieves very high precision and MS at the same time.


\subsection{Feature Matching in Independent Motion}
One may wonder if the proposed relative geometric constraint may cause problem when there exist independent motions.
Although the relative geometry between features in the same object will remain similar, the features from different objects would not have such geometric consistency. 
In the proposed algorithm, 
if an edge is on the motion boundary, the mutual contribution of the nodes to the other nodes becomes very weak due to the motion inconsistency. 
This in effect cuts the edges on the motion boundaries, thus only the features with similar motion contributes to the estimation of candidate match score.

The proposed algorithm is tested with the images of public dataset \cite{wong2011dynamic}, and two representative results are displayed in Figure~\ref{fig:independent_motion} (more results are in the supplementary material). 
From the experimental results it is shown that the proposed algorithm can handle the independent motions. 


\begin{figure}
\centering
\subfloat{\includegraphics[trim=0.4cm 0.4cm 0.4cm 0.4cm, width=0.95\linewidth]{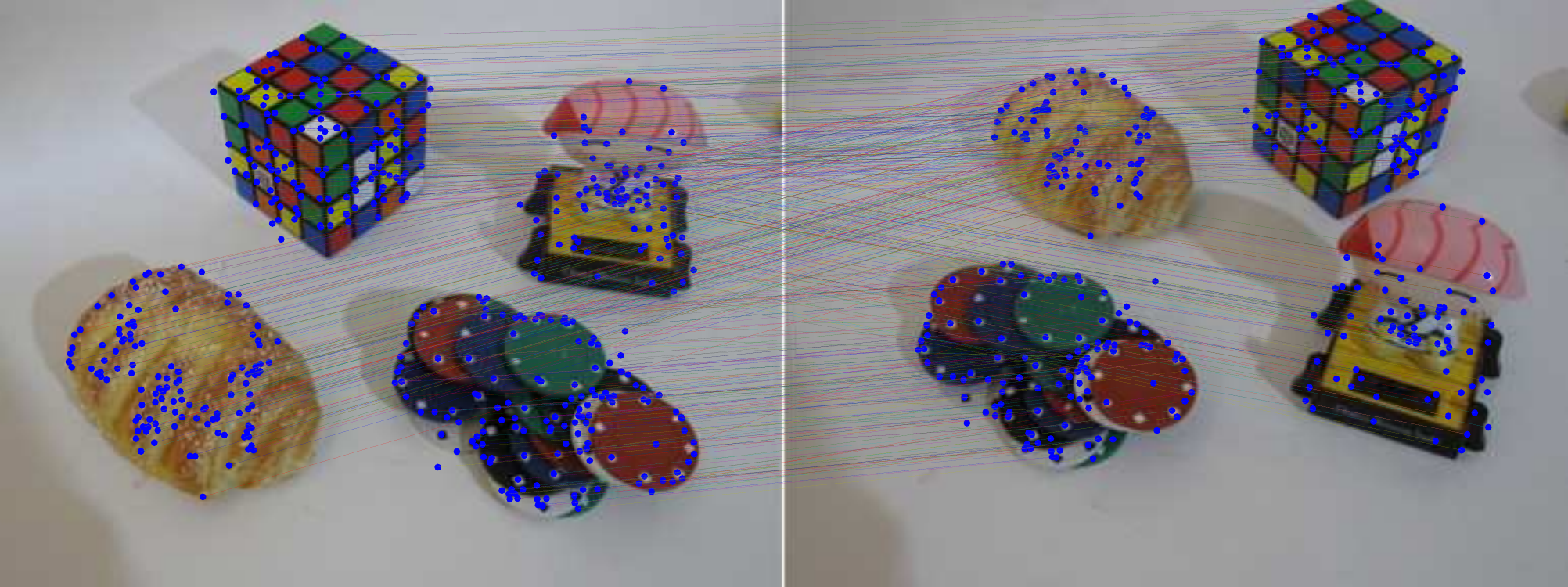}} \\
\subfloat{\includegraphics[trim=0.4cm 0.4cm 0.4cm 0.4cm, width=0.95\linewidth]{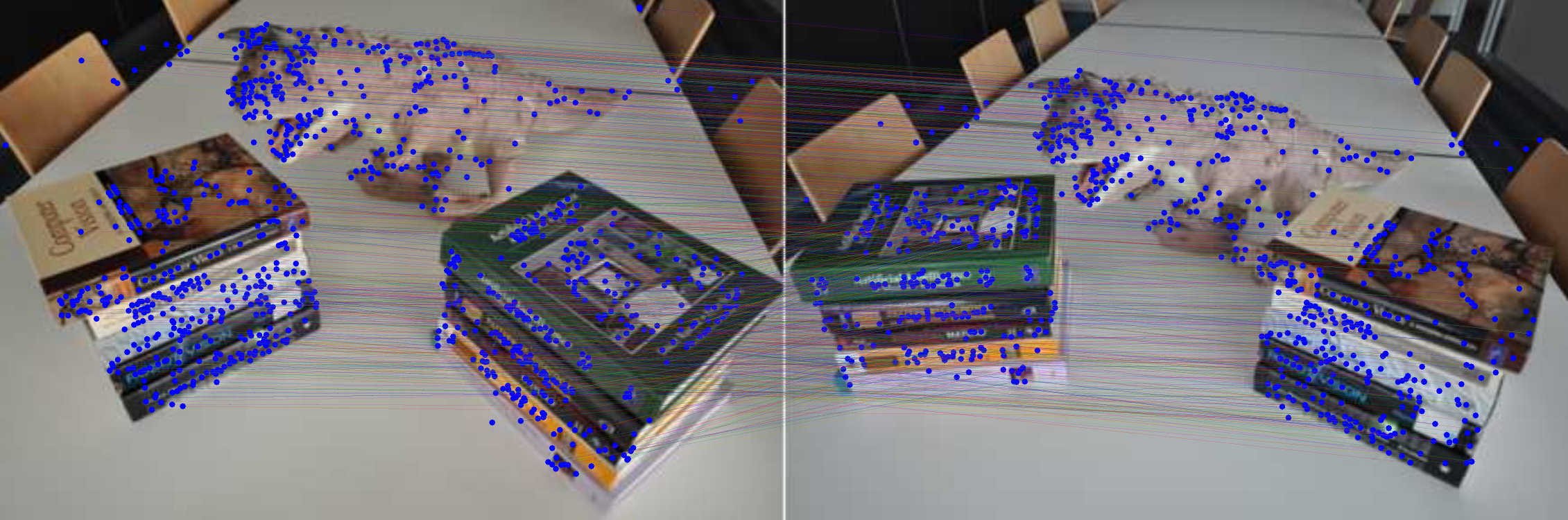}} \\
\caption{ \small
The results on feature matching in the presence of independent motions. 
The MRF edges across motion boundary are effectly disabled automatically by the algorithm.
They are by feature matching only without any post processing. 
}
\vspace{-0.3cm}
\label{fig:independent_motion}
\end{figure}

\section{Conclusion and Future Work}
In this paper, we propose a feature matching algorithm that utilizes both geometric properties and descriptor of local features.
The feature matching problem is formulated as an MRF model, and the solution is found by the min-sum BP algorithm.
Bidirectional transfer measure computes the relative geometric consistency of a pair of feature correspondences, and it is used to filter the similar-looking match candidates in the target image.
Progressive correspondence search is introduced to find the solution efficiently and robustly.
Through extensive experiments, we quantitatively evaluated that the proposed method outperforms both graph-based and descriptor-based conventional feature matching algorithms.


{\small
\bibliographystyle{ieee}
\bibliography{cvpr16_feature_match}
}

\end{document}